\pdfoutput=1
\documentclass[conference]{IEEEtran}

\usepackage{tabularx}
\usepackage{booktabs}

\usepackage[utf8]{inputenc}
\usepackage[T1]{fontenc}
\usepackage{microtype}

\usepackage{hyperref}
\usepackage{url}

\usepackage{graphicx}
\usepackage{subfigure}          % or replace with subcaption if preferred
\usepackage{booktabs}
\usepackage{multirow}
\usepackage{adjustbox}

\usepackage{amsmath}
\usepackage{amssymb}
\usepackage{siunitx}

\usepackage{xspace}
\usepackage{paralist}
\usepackage{enumitem}
\usepackage{verbatim}
\usepackage[hang,flushmargin]{footmisc}

\usepackage[table,xcdraw]{xcolor}
\usepackage{csquotes}
\usepackage{latexsym}

\title{ROMEVA: Geometry-Preserving Vocabulary Expansion for Roman Urdu Language Models}
\author{

\IEEEauthorblockN{
Mahnoor Khan$^{1,*}$,
Afsheen Asif$^{1,*}$,
Milhan Afzal Khan$^{2}$,
Seemab Latif$^{1}$,
Mehwish Fatima$^{1,\dagger}$
}

\IEEEauthorblockA{
$^{1}$School of Electrical Engineering and Computer Science (SEECS), \\
National University of Sciences and Technology (NUST), \\ 
Islamabad, Pakistan\\
\{mkhan.msds25seecs, aasif.msds25seecs, seemab.latif, mehwish.fatima\}@seecs.edu.pk
}

\IEEEauthorblockA{
$^{2}$Department of Computer Science, \\
University of Agriculture Faisalabad, \\
Faisalabad, Pakistan\\
milhankhan@uaf.edu.pk
}

\IEEEauthorblockA{
$^{*}$Equal contribution \\
$^{\dagger}$Corresponding author: mehwish.fatima@seecs.edu.pk
}

}

\begin{document}
\maketitle
\begin{abstract}
Multilingual Language Models like mBERT are widely used for low-resource NLP, yet their adaptation to morphologically inconsistent languages such as Roman Urdu remains underexplored. Roman Urdu spelling variation causes severe sub-word fragmentation, averaging 1.50 sub-words per token. We propose \textit{ROMEVA} (Roman Urdu Embedding-preserving Vocabulary Adaptation), which combines sub-word-average initialization and a PCA-guided anchor loss to stabilize embeddings during vocabulary expansion. Using a 36,130-comment Roman Urdu corpus, we add 500 highly fragmented tokens to mBERT and compare naive fine-tuning, sub-word-aware fine-tuning, and \textit{ROMEVA}. While \textit{ROMEVA} most effectively preserves the pretrained embedding space, naive fine-tuning achieves the strongest downstream sentiment classification performance. These findings reveal a disconnect between embedding stability and downstream performance, suggesting that stronger adaptation may be preferable to strict embedding preservation in morphologically inconsistent languages.
\end{abstract}

%=============================================================
\section{Introduction}
Multilingual pretrained language models such as mBERT \cite{devlin-etal-2019-bert} have become pivotal to low-resource NLP because they transfer knowledge across languages through shared sub-word representations. These models perform well when the tokenizer aligns with the target language, but struggles with orthographically inconsistent languages. Table~\ref{tab:intro_example} illustrates the challenge. Semantically identical Roman Urdu words often appear in multiple spellings, each producing different sub-word segmentation under mBERT. As a result, the model distributes evidence for the same concept across multiple fragmented representations, reducing vocabulary coverage and weakening lexical consistency. Such variation fragments semantically related words into multiple sub-words and weakens the quality of representations. Roman Urdu is the dominant language of digital communication for millions of users across Pakistan and South Asia, particularly on social media, messaging platforms, and online forums\cite{bilal2023roman}. Its effective modeling is therefore important for downstream applications such as sentiment analysis, hate speech detection, information retrieval, and conversational AI.

A common response to vocabulary mismatch is domain-adaptive pretraining (DAPT) \cite{gururangan-etal-2020-dont} and vocabulary expansion, where domain-specific tokens are added to a pretrained model and the model is further adapted to target-domain data. These approaches improve vocabulary coverage and reduce fragmentation without requiring pretraining from scratch. However, multilingual tokenizers such as BPE \cite{sennrich-etal-2016-neural} and SentencePiece \cite{kudo-richardson-2018-sentencepiece} assume relatively stable surface forms, an assumption often violated in Roman Urdu due to extensive spelling variation. While prior work has explored Roman Urdu resources and downstream applications \cite{rizwan2020hate,nazir2025leveraging,soomro2024dataset}, multilingual tokenization \cite{devlin-etal-2019-bert}, vocabulary adaptation \cite{gururangan-etal-2020-dont}, and representation stability \cite{kirkpatrick2017overcoming,mu-viswanath-2018-all,ethayarajh-2019-contextual} independently, the interaction between vocabulary expansion and embedding drift remains largely unexplored. In particular, no previous work has systematically investigated how vocabulary expansion affects pretrained embedding geometry in Roman Urdu \cite{gururangan-etal-2020-dont,kirkpatrick2017overcoming,mu-viswanath-2018-all} or whether preserving that geometry benefits downstream task performance.

\begin{table}[t]
\centering
\caption{Examples of Roman Urdu spelling variation and mBERT tokenization.}
\label{tab:intro_example}
\footnotesize
\begin{tabular}{lll}
\toprule
\textbf{Meaning} & \textbf{Roman Urdu Forms} & \textbf{mBERT Tokens} \\
\midrule
What &
kya, kia, kiya &
ky + \#\#a,\;
ki + \#\#a,\;
ki + \#\#ya \\
%\midrule
Very &
bohot, bohat &
bo + \#\#hot,\;
bo + \#\#hat \\
%\midrule
Need &
chahiye &
cha + \#\#hi + \#\#ye \\
%\midrule
Sometimes &
kabhi, kbhi &
ka + \#\#bhi,\;
kb + \#\#hi \\
\bottomrule
\end{tabular}
\end{table}

To address this gap, we investigate the relationship between vocabulary expansion, embedding stability, and downstream performance in Roman Urdu. We introduce \textit{ROMEVA} (\textit{Roman Urdu Embedding-preserving Vocabulary Adaptation}), a framework that combines sub-word-average initialization with a PCA-guided anchor loss to reduce embedding drift during vocabulary adaptation. Using a newly collected corpus of 36,130 Roman Urdu YouTube comments, we identify highly fragmented tokens, expand the mBERT vocabulary, and compare naive vocabulary adaptation, sub-word-aware initialization, and \textit{ROMEVA}. 

We summarize our contributions as follows:
\begin{itemize}
     \item We introduce \textit{RUVA}, a Roman Urdu datasey comprising 36,130 YouTube comments, designed for studying tokenization behavior and vocabulary adaptation in multilingual language models.
    \item We provide the first systematic analysis of tokenization-induced embedding drift in Roman Urdu vocabulary adaptation.
    \item We propose \textit{ROMEVA}, a geometry-preserving vocabulary expansion framework for adapting pretrained language models to Roman Urdu.
    \item We demonstrate that improved embedding stability does not necessarily translate into better downstream sentiment classification performance, revealing an important trade-off between representation preservation and task-specific adaptation.
\end{itemize}

%Our study makes three contributions: (1) we provide the first systematic analysis of tokenization-induced embedding drift in Roman Urdu vocabulary adaptation; (2) we propose ROMEVA, a geometry-preserving vocabulary expansion framework; and (3) we demonstrate that improved embedding stability does not necessarily translate into better downstream sentiment classification performance, revealing an important trade-off between representation preservation and task-specific adaptation.

%=============================================================
\section{Related Work}
We divide this section into three directions: datasets for Roman Urdu, vocabulary generation techniques, and preservation of embedding geometry.  

\subsection{Roman Urdu Resources}
The scarcity of annotated Roman Urdu corpora remains a bottleneck in developing robust NLP systems. \cite{shafi2023unlt} develops an Urdu Natural Language Toolkit (UNLT) that consists of a word tokenizer, a sentence tokenizer, and a POS tagger. Although UNLT targets native Urdu script, it highlights a challenge of Roman Urdu: morphologically rich text with inconsistent word boundaries is difficult for standard tokenization approaches\cite{shafi2023unlt}. Early Roman Urdu datasets mainly focus on classification tasks under limited-resource settings. \cite{rizwan2020hate} introduce RUHSOLD, a corpus of 10,012 Roman Urdu tweets for hate speech and offensive language detection. They show the effectiveness of transfer learning for Roman Urdu in these downstream tasks\cite{rizwan2020hate}. Similarly, code-mixed datasets comibining Roman-Urdu and -Punjabi \cite{nazir2025leveraging} demonstrate the effectiveness of transformer-based models on informal social media text. However, tokenization issues and vocabulary coverage remain unexplored. \cite{soomro2024dataset} introduce Roman Urdu Word Variations and Normalized Sentiment Review dataset (RUWV-NSR) \cite{soomro2024dataset}, containing 28,090 Roman Urdu reviews annotated into five sentiment classes. These resources highlight the prevalence of orthographic inconsistency in Roman Urdu but do not investigate its impact on tokenization and vocabulary adaptation. 

\subsection{Vocabulary Adaptation}
Modern multilingual language models rely on sub-word tokenization methods such as Byte Pair Encoding (BPE) \cite{sennrich-etal-2016-neural} and SentencePiece \cite{kudo-richardson-2018-sentencepiece}. While effective for standardized languages, these approaches often struggle with informal and romanized text where spelling variation is common. Multilingual BERT (mBERT) \cite{devlin-etal-2019-bert}, trained primarily on formal Wikipedia text from 104 languages, does not explicitly model Roman Urdu and therefore frequently fragments Roman Urdu words into multiple sub-word units. A common strategy for addressing vocabulary mismatch is domain-adaptive pretraining (DAPT) \cite{gururangan-etal-2020-dont} or vocabulary expansion, where domain-specific tokens are added to the pre-trained vocabulary and the model is further adapted to target-domain data. Although these approaches improve vocabulary coverage, existing work largely focuses on adaptation effectiveness and assumes the pretrained embedding space remains stable throughout the process.

\subsection{Representation Stability}
Representation stability has been studied extensively in continual learning and embedding geometry research. Elastic Weight Consolidation (EWC) \cite{kirkpatrick2017overcoming} mitigates catastrophic forgetting by constraining changes to important parameters during adaptation. At the representation level, \cite{mu-viswanath-2018-all} show that semantic information is concentrated in a small number of principal directions, while \cite{ethayarajh-2019-contextual} demonstrate that contextualized embeddings exhibit structured geometric properties. These findings suggest that preserving important geometric directions may help maintain representation quality during adaptation. Recent work has also explored adaptive tokenization strategies. For example, \cite{wang2025pit} jointly optimize tokenization and representation learning in recommendation systems, demonstrating the benefits of adapting tokenization to the target domain. However, prior work has not examined how vocabulary expansion affects pretrained embedding geometry in morphologically inconsistent languages such as Roman Urdu.

%=============================================================
\begin{figure}[t]
    \centering
    \includegraphics[width=\columnwidth]{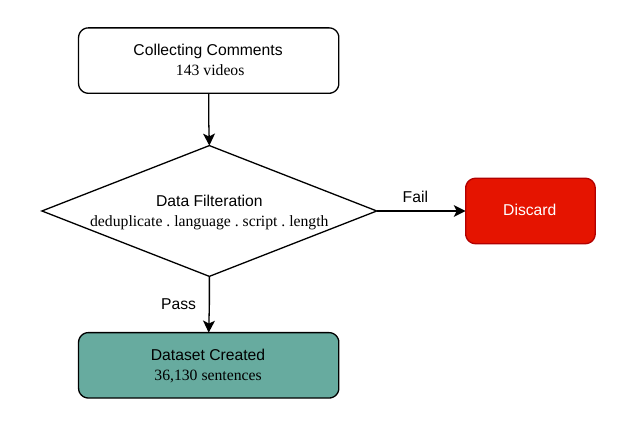}
    \caption{Dataset constuction.}
    \label{fig:workflow}
\end{figure}

\section{RUVA: Roman Urdu Vocabulary Adaptation Dataset}
To investigate vocabulary adaptation in Roman Urdu, we construct \textbf{RUVA}, a corpus of Roman Urdu YouTube comments collected from Pakistani drama and podcast channels. The dataset serves two purposes. First, it provides a large sample of naturally occurring Roman Urdu for analyzing tokenization behavior in multilingual language models. Second, it enables the identification of vocabulary items that are systematically fragmented by mBERT and therefore candidates for vocabulary expansion.

\subsection{Construction}
We collect comments from 143 YouTube videos using the YouTube Data API v3. The corpus undergoes a sequence of filtering steps to retain Roman Urdu text while removing noise. Exact-match comparison removes duplicate comments. An English lexicon filter removes comments in which more than 85\% of the tokens correspond to English words. A Unicode range filter (\texttt{[\textbackslash u0600-\textbackslash u06FF]}) excludes comments containing native Urdu script. Finally, we remove comments shorter than five whitespace-tokenized words to eliminate low-information responses such as emojis, acknowledgements, and isolated tokens. Figure~\ref{fig:workflow} summarizes the dataset construction pipeline. The final corpus contains 36,130 Roman Urdu comments.

\begin{figure*}[h]
    \centering
    \includegraphics[width=\textwidth]{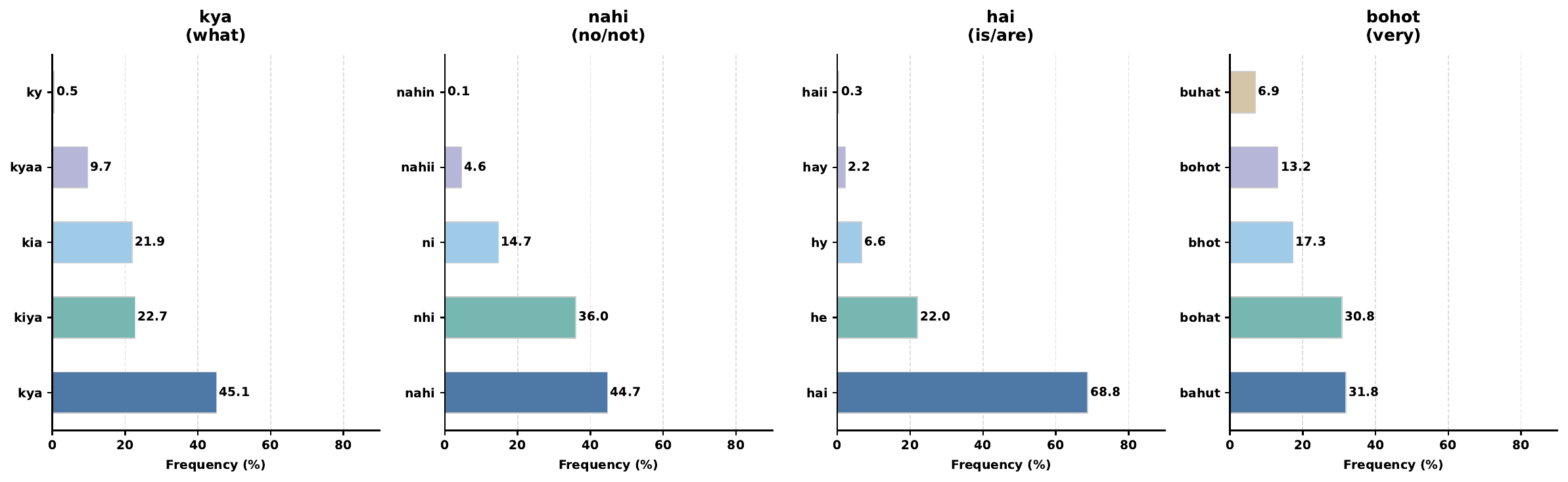}
    \caption{Spelling variation distribution for high-frequency Roman Urdu words.}
    \label{fig:spelling_variants}
\end{figure*}

\subsection{Analysis}
The final dataset contains 36,130 comments, 505,314 word tokens, and a vocabulary of 38,250 unique forms. Table~\ref{tab:1} summarizes the corpus characteristics and the statistics used in subsequent analyses. Two properties are particularly relevant for vocabulary adaptation: extensive orthographic variation and severe sub-word fragmentation.

\begin{table}[t]
\caption{Summary statistics of the Roman Urdu corpus.}
\label{tab:1}
\centering
\renewcommand{\arraystretch}{1.2}
\begin{tabular}{ll}
\toprule
\textbf{Statistic} & \textbf{Value} \\
\midrule
Total comments & 36,130\\
Total word tokens & 505,314\\
Unique vocabulary & 38,250 \\
Average sentence length & 13.99 words\\
Median sentence length & 9 words\\
Hapax legomena & 23,061 (60.3\%) \\
Words fragmented (2+sub-words) & 37.9\% \\
Mean fragmentation ratio & 1.50 sub-words/word\\
OOV candidates (3+sub-words) & 14,962 \\
Valid OOV candidates & 14,206\\
Selected expansion tokens & 500 \\
%Missing values & 0 \
%Duplicates & 0 \
\bottomrule
\end{tabular}
\end{table}

\subsubsection{Orthographic Variation}
Orthographic variation is a defining characteristic of Roman Urdu. Of the 38,250 unique vocabulary items, 23,061 (60.3\%) occur only once, indicating substantial lexical sparsity. A large portion of this sparsity arises because alternative spellings of the same word appear as separate vocabulary entries. Figure~\ref{fig:spelling_variants} shows this phenomenon for several high-frequency words. The word \texttt{hai (``is/are'')} appears in multiple forms including \texttt{hai}, \texttt{he}, and \texttt{hy}, while \texttt{kya (``what'')} occurs as \texttt{kya}, \texttt{kia}, and \texttt{kiya}. Similar variation appears throughout the corpus, substantially increasing the effective vocabulary size despite a smaller underlying semantic inventory. This orthographic inconsistency creates a challenging setting for multilingual tokenizers.

\subsubsection{Tokenization Mismatch}
To quantify the impact of orthographic variation on multilingual tokenization, we apply the mBERT tokenizer to the entire corpus. 

\begin{figure}[ht]
    \centering
    \includegraphics[width=\linewidth]{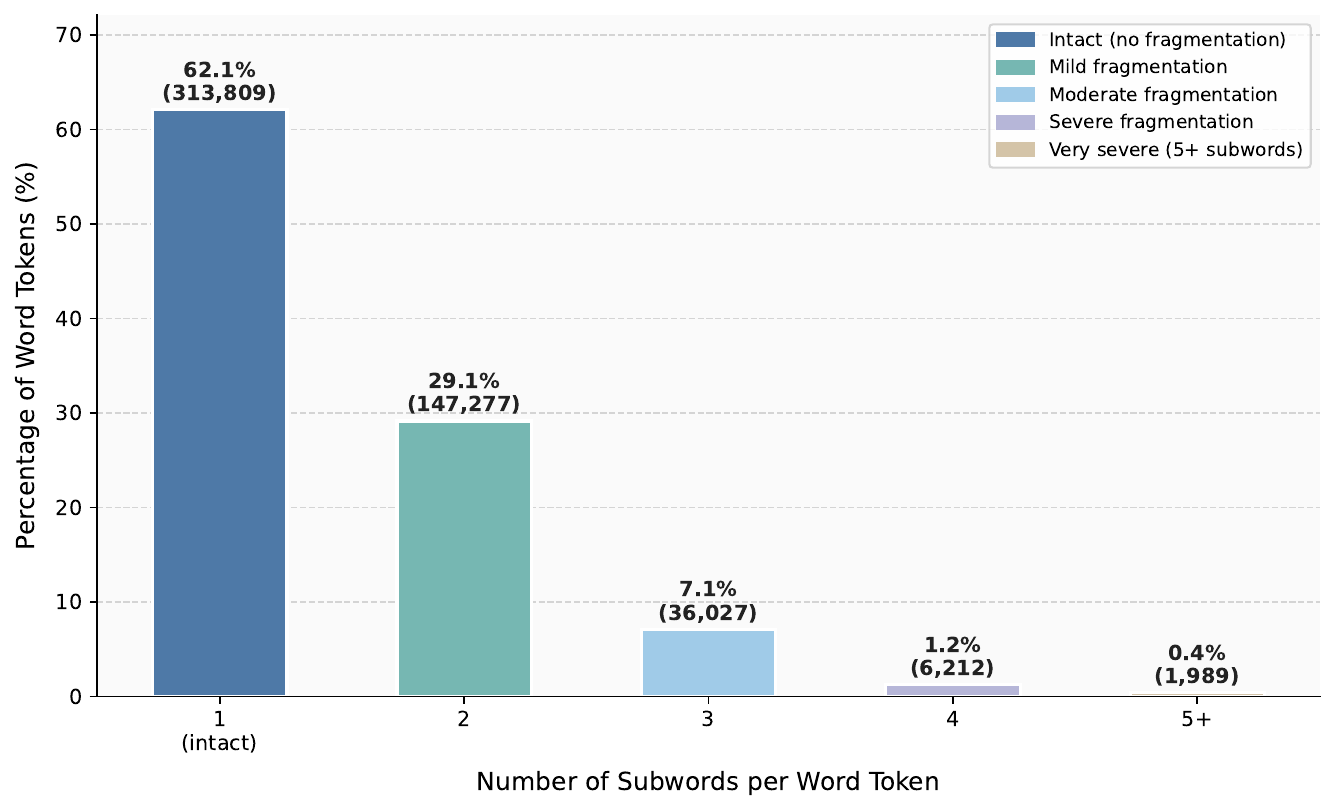}
    \caption{mBERT sub-word fragmentation distribution across 505,314 Roman Urdu word tokens (mean = 1.50 sub-words/word).}
    \label{fig:frag_dist}
\end{figure}

Figure~\ref{fig:frag_dist} shows that 37.9\% of the words fragment into multiple sub-words. While 8.7\% split into 3 or more pieces. The Average fragmentation ratio is 1.5 sub-words per word. These results show that there is a considerable mismatch between the mBERT's original vocabulary and the naturally written Roman Urdu vocabulary. Many common words of Roman Urdu are represented in the form of fragmented sub-word sequences instead of a single lexical unit, highlighting the need for vocabulary adaptation.

\subsubsection{Vocabulary Expansion Candidates}
We identify those tokens for vocabulary expansion that split into three or more words during mBERT tokenization. 14,962 candidates are identified in this process. 14,204 characters remained after the filtering process, including alphabetic content, length
between 3 and 20 characters, and excessive character repetition. 
The resulting list contains frequently appearing Roman urdu words such as \textit{chahiye}, \textit{kbhi} , and \textit{Pakistani} in the corpus, but mBERT consistently fragments them. We rank the candidates by frequency and select top 500 tokens for vocabulary expansion. This increases the mBERT vocabulary from 119,547 to 120,047 tokens.

\subsection{Candidate Selection}
\begin{figure}[t]
    \centering
    \includegraphics[width=\linewidth]{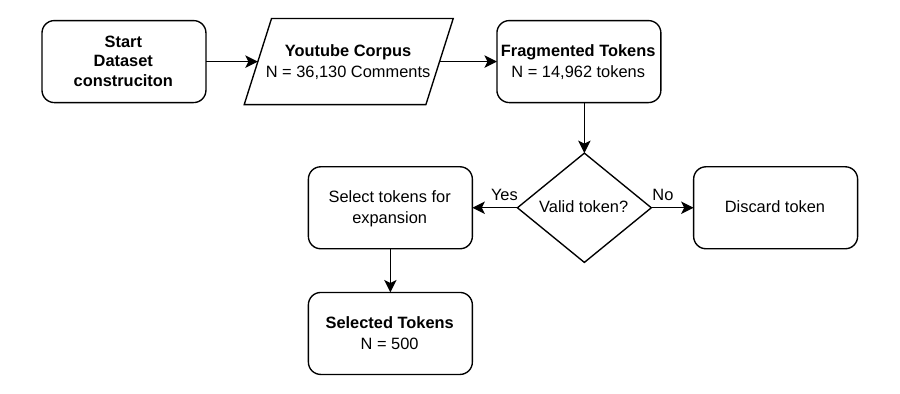}
    \caption{Vocabulary expansion candidate selection pipeline.}
    \label{fig:oov_pipeline}
\end{figure}

Vocabulary expansion requires identifying Roman Urdu words that are poorly represented by the pretrained tokenizer. We analyze RUVA to identify out-of-vocabulary (OOV) candidates. We filter tokens using frequency thresholds and tokenization constraints, and then rank valid candidates by corpus frequency. The resulting vocabulary items serve as the basis for vocabulary expansion in mBERT. Figure~\ref{fig:oov_pipeline} summarizes the candidate selection process.

%--------------------------------------------
\section{ROMEVA: Roman Urdu Embedding-preserving Vocabulary Adaptation}
\begin{figure}[ht]
    \centering
    \includegraphics[width=0.3\textwidth]{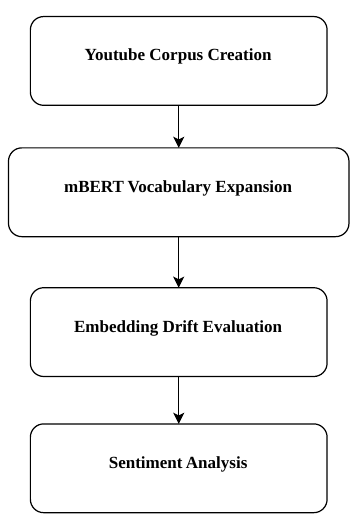}
    \caption{ROMEVA pipeline overview}
    \label{fig:pipeline}
\end{figure}

Vocabulary expansion improves coverage for fragmented Roman Urdu words but introduces two challenges. Newly added tokens require meaningful initialization, and continued pretraining can distort pretrained embeddings. \textit{ROMEVA} addresses both issues through sub-word-average initialization and a PCA-guided anchor loss. Figure~\ref{fig:pipeline} presents the overall framework. 

Fig \ref{fig:pipeline} illustrates the end-to-end experimental pipeline. We first create the YouTube corpus, then we expand the vocabulary of mBERT using the tokens from this corpus. We calculate the embedding drift on the vanilla mBERT, with its embeddings
initialized as sub-word averages and the similar approach with anchor loss. Later, we carry out sentiment analysis of the models.

\subsection{Sub-word Average Initialization}
Vocabulary expansion typically initializes new tokens randomly. Such embeddings lack semantic grounding and require substantial adaptation during continued pretraining, increasing pressure on the pretrained embedding space. \textit{ROMEVA} initializes each new token using the average of its constituent sub-word embeddings. Given a token \(t\) segmented into \(n\) existing sub-words \(\{s_1,\ldots,s_n\}\), its initial embedding is:

{\small
\[
E_t = \frac{1}{n}\sum_{i=1}^{n} E_{s_i}.
\]
}

For example, the Roman Urdu token "Karwayein" is broken into "\texttt{\#\#kar}," "\texttt{\#\#wa}," "\texttt{\#\#ye}," and "\texttt{\#\#in}." The average of these four existing sub-word embeddings is calculated and used as the starting representation for "Karwayein." The new token starts fine-tuning from a more informed position rather than a random one since mBERT already has meaningful representations for these sub-words from pretraining on multilingual data.

\subsection{PCA-Guided Anchor Loss}  
Even with improved initialization, continued pretraining can shift pretrained embeddings away from their original geometry. \textit{ROMEVA} mitigates this drift through an anchor loss that preserves variation along the most informative embedding directions. Before adaptation, we perform principal component analysis on the pretrained mBERT embedding matrix and retain the top 128 principal components, which explain 37.5\% of the embedding variance. During training, deviations from the pretrained embeddings along these directions are penalized:

{\small
\[
L_{\text{total}} = L_{\text{MLM}} + \frac{\lambda}{N}
\sum_{i=1}^{N}
\left\| \operatorname{proj}(E_{\text{cur}} - E_{\text{orig}}, C) \right\|^2
\]}

where, \(E_{\text{orig}}\) and \(E_{\text{cur}}\) denote the pretrained and current embedding matrices, \(C\) contains the top principal components, and \(\lambda\) controls the regularization strength. This formulation preserves semantically important directions while allowing adaptation in lower-variance dimensions. Overall, sub-word-average initialization and anchor regularization balance vocabulary adaptation with preservation of pretrained knowledge.

%=======================================================================
\section{Experimental Design}
\subsection{Datasets}
We use two datasets in our experiments. Vocabulary adaptation is performed on \textbf{RUVA}, our Roman Urdu Vocabulary Adaptation Dataset comprising 36,130 YouTube comments. Downstream evaluation uses the RUWV-NSR sentiment dataset \cite{soomro2024dataset}. From RUVA, we identify 14,206 valid out-of-vocabulary (OOV) candidates and select the 500 most frequent tokens for vocabulary expansion. These tokens are added to the mBERT vocabulary, increasing its size from 119,547 to 120,047 entries. Embedding drift is subsequently analyzed on 9,013 corpus-relevant tokens. For sentiment classification, we adopt the standard 80/20 train--test split of RUWV-NSR, resulting in 21,457 training instances and 5,365 test instances.

\subsection{Base Model}
We use multilingual BERT (mBERT) \cite{devlin-etal-2019-bert} as the base model. mBERT contains 110M parameters and is pretrained on Wikipedia text from 104 languages using masked language modeling and next sentence prediction objectives. We select mBERT because it is the most widely used multilingual encoder for Roman Urdu and exhibits substantial tokenization mismatch on our corpus, making it an appropriate platform for studying vocabulary adaptation. All experimental conditions start from the same pretrained checkpoint.

\subsection{Baselines}
We compare \textit{ROMEVA} against two vocabulary adaptation baselines.

\subsubsection{Naive Expansion} extends the mBERT vocabulary with the selected Roman Urdu tokens and initializes the corresponding embeddings randomly. The model is then adapted through continued masked language model training without any additional regularization.
\subsubsection{Sub-word Initialization} uses the same expanded vocabulary but initializes each new token with the average of its constituent mBERT sub-word embeddings.% Unlike ROMEVA, no geometry-preserving regularization is applied during adaptation. This baseline isolates the effect of informed embedding initialization from that of anchor-based embedding preservation.

\subsection{Evaluation Metrics}
We evaluate both embedding stability and downstream task performance.

\subsubsection{Embedding Drift}
We measure drift after MLM adaptation using: (i) \textit{L2 Drift (L2D)}, the mean Euclidean distance between pretrained and adapted embeddings; and (ii) \textit{Cosine Similarity (CoSim)}, the directional similarity between pretrained and adapted embeddings. Both metrics are computed over corpus-relevant tokens as well as the full vocabulary.

\subsubsection{Downstream Performance}
We evaluate sentiment classification on RUWV-NSR using Accuracy, Macro F1, and Weighted F1.

%=========================================================
\section{Results}

\subsection{Embedding Stability}
Table~\ref{tab:drift} compares embedding drift across the three adaptation strategies. Both sub-word-aware initialization and anchor regularization substantially reduce drift relative to naive adaptation. \textit{ROMEVA} achieves the lowest L2 drift and highest cosine similarity, indicating the strongest preservation of the pretrained embedding space.

The improvement is particularly evident for corpus-relevant tokens, where adaptation effects are concentrated. In contrast, drift measured over the full vocabulary is influenced by many tokens that never appear in the Roman Urdu corpus and therefore remain largely unchanged. These results suggest that sub-word-aware initialization provides a stable starting point for newly introduced tokens, while the anchor loss further constrains movement along semantically important directions.

\begin{table}[ht]
\caption{Embedding drift results across experimental conditions.}
\label{tab:drift}
\centering
\begin{tabular}{lccc}
\toprule
\textbf{Metrics} & \textbf{Naive} & \textbf{Sub-word} & \textbf{ROMEVA} \\
\midrule
L2D (Relevant) $\downarrow$ & 31.89 & 27.09 & \textbf{26.90} \\
L2D (All) $\downarrow$      & 44.91 & 23.81 & \textbf{23.48} \\
Cosine Sim $\uparrow$         & 96.01 & 97.04 & \textbf{97.08} \\
\bottomrule
\end{tabular}

\end{table}

%When measuring drift across all 119,547 tokens, the values decrease significantly: 0.4491 for Naive, 0.2381 for Sub-word (No Anchor), and 0.2348 for ROMEVA. This reinforces the earlier choice to limit evaluation to corpus-relevant tokens. The full-vocabulary averages are influenced by the large number of tokens that do not appear in the Roman Urdu training data and thus remain unchanged, which obscures the true extent of drift in the important areas of the embedding space.

%Cosine similarity scores support the L2 findings. The Naive condition produces the lowest cosine similarity of 0.9601, while Sub-word (No Anchor) and ROMEVA improve to 0.9704 and 0.9708 respectively. This indicates that the anchor-regularized and sub-word-aware conditions better maintain the directional structure of the pretrained embeddings.

\subsection{Downstream Sentiment Classification}
Table~\ref{tab:sentiment} presents sentiment classification results on RUWV-NSR. Contrary to the embedding-level findings, naive fine-tuning achieves the strongest performance across all evaluation metrics. Both sub-word-aware initialization and \textit{ROMEVA} lead to progressively lower accuracy and F1 scores.

These results indicate that improved embedding stability does not necessarily translate into better downstream performance. While \textit{ROMEVA} preserves the pretrained embedding space more effectively, the corresponding constraints appear to limit adaptation to Roman Urdu-specific lexical and semantic patterns. In this setting, larger representational shifts produced by unconstrained fine-tuning yield superior task-level performance.

\begin{table}[ht]
\caption{Sentiment classification results on the RUWV-NSR benchmark.} %Sentiment classification results on the Mendeley test set.}
\label{tab:sentiment}
\centering
\begin{tabular}{lccc}
\toprule
\textbf{Metrics} & \textbf{Naive} & \textbf{Sub-word} & \textbf{ROMEVA} \\
\midrule
Accuracy $\uparrow$    & \textbf{63.3} & 61.7 & 60.9 \\
Macro F1 $\uparrow$    & \textbf{64.5} & 63.0 & 62.3 \\
Weighted F1 $\uparrow$  & \textbf{63.0} & 62.0 & 61.0 \\
\bottomrule
\end{tabular}

\end{table}

\subsection{Class-wise Performance}
Table~\ref{tab:perclass} presents class-wise F1 scores for the three adaptation strategies. Performance varies considerably across sentiment categories. Very Positive and Very Negative reviews obtain highest F1 score across all models, suggesting that strong sentiment cues are relatively easy to identify. Although, the remaining classes are more challenging, probably because the lexical signals overlap and the decision boundaries are not clear enough. 
The relative ranking of adaptation strategies between classes remains fairly consistent. Though, \textit{ROMEVA} stabilizes embeddings, it does not yield better class-level classification performance. This observation compacts the finding that preserving pretrained representations and optimizing downstream task performance are not necessarily aligned objectives. 

\begin{table}[ht]
\caption{Class-wise F1 scores on the RUWV-NSR benchmark. Higher values indicate better performance.}
\label{tab:perclass}
\centering
\begin{tabular}{lccc}
\toprule
\textbf{Class} & \textbf{Naive} & \textbf{Sub-word} & \textbf{ROMEVA} \\
\midrule
Very Negative & 75.0 & \textbf{77.0} & 74.0 \\
Negative & \textbf{50.0} & 47.0 & 46.0 \\
Neutral & \textbf{62.0} & 60.0 & 60.0 \\
Positive & 52.0 & \textbf{53.0} & 52.0 \\
Very Positive & \textbf{81.0} & \textbf{81.0} & \textbf{81.0} \\
\bottomrule
\end{tabular}

\end{table}

\subsection{Stability--Performance Trade-off}
Overall, the results reveal a clear tradeoff between the embedding stability and the downstream adaptation. \textit{ROMEVA} consistently achieves lowest embedding drift and highest cosine similarity, which shows that \textit{ROMEVA} preserves the embedding geometry. But, these gains do not benefit sentiment classification performance. Instead, the naive approach performs better on all task-level metrics.
These findings suggest that preserving pretrained representations and optimizing downstream task performance do not necessarily align in the same direction. The mismatch in the morphologically inconsistent languages like Roman Urdu can be so huge that it might require larger modifications than regularized adaptations in the embedding space. This is why the benefits of aggressive adaptation for sentiment classification task outweigh the benefits of strict embedding preservation.
In general, the results highlight the need to evaluate vocabulary adaptation methods beyond embedding-level metrics alone. Improvements in representation stability do not automatically imply gains in downstream performance, and future work should explicitly consider the trade-off between preserving pretrained knowledge and adapting to target-domain linguistic variation.

\section{Limitations}
Our study has the following limitations. First, we evaluate \textit{ROMEVA} on Roman Urdu only, so it remains unclear it remains unclear whether this trade-off generalizes to other morphologically inconsistent or romanized languages. Second, we conduct our experiments using mBERT only. So, it will be an interesting study that how different multilingual architectures and tokenization schemes respond to vocabulary expansion and geometry-preserving regularization. Third, vocabulary expansion is limited to the 500 most frequent candidate tokens. The effect of larger vocabulary additions and alternative token selection strategies remains unexplored. Fourth, downstream evaluation is performed to sentiment classification on RUWV-NSR. Other tasks might exhibit different trade-offs between representation preservation and adaptation. Finally, embedding stability is measured using L2 drift and cosine similarity, which capture changes in embedding geometry but might not fully reflect effects on contextual representations and higher-layer model behavior.

\section{Conclusions}
Roman Urdu poses a significant challenge for multilingual language models because orthographic variation creates vocabulary mismatch and severe sub-word fragmentation. We investigate this problem through a newly constructed Roman Urdu YouTube corpus and introduce \textit{ROMEVA}, a vocabulary adaptation framework that combines sub-word-aware initialization with geometry-preserving regularization. \textit{ROMEVA} effectively reduces embedding drift during vocabulary expansion, but improved embedding stability does not translate into better downstream performance. Instead, unconstrained adaptation achieves the strongest sentiment classification results. These findings suggest that, for morphologically inconsistent languages such as Roman Urdu, the benefits of adaptation may outweigh the benefits of strict embedding preservation.

\bibliographystyle{IEEEtran}
\bibliography{references}

% \appendix

% \section{Example Appendix}
% \label{sec:appendix}
% This is a simple appendix section as a placeholder. Use \verb|\appendix| before starting appendices.

\end{document}